%% file: main.tex
\documentclass{article}

\usepackage{arxiv}

\usepackage[utf8]{inputenc} 
\usepackage[T1]{fontenc}    
\usepackage{hyperref}       
\usepackage{url}            
\usepackage{booktabs}       
\usepackage{amsfonts}       
\usepackage{nicefrac}       
\usepackage{microtype}      
\usepackage{lipsum}         
\usepackage{graphicx}
\usepackage{natbib}
\usepackage{doi}
\usepackage{amsmath,amssymb,amsfonts,dsfont}
\usepackage[table]{xcolor}
\usepackage{multirow}
\usepackage{algorithm}
\usepackage{algpseudocode}
\usepackage{cleveref}       

\setcitestyle{numbers,square}

\title{Data-Driven Diagnosis for Large Cyber-Physical-Systems with Minimal Prior Information}

\newif\ifuniqueAffiliation

\ifuniqueAffiliation 
\author{Henrik Sebastian Steude \\
	Institute of Automation Technology\\
	Helmut Schmidt University\\
	Holstenhofweg 85, 22043 Hamburg, Germany \\
	\texttt{henrik.steude@hsu-hh.de} \\
	\And
	Alexander Diedrich \\
	Institute of Automation Technology\\
	Helmut Schmidt University\\
	Holstenhofweg 85, 22043 Hamburg, Germany \\
	\texttt{alexander.diedrich@hsu-hh.de} \\
	\And
	Ingo Pill \\
	Institute of Software Technology\\
	Graz University of Technology\\
	Rechbauerstraße 12, 8010 Graz, Austria \\
	\texttt{ingo.pill@tugraz.at} \\
	\And
	Lukas Moddemann \\
	Institute of Automation Technology\\
	Helmut Schmidt University\\
	Holstenhofweg 85, 22043 Hamburg, Germany \\
	\texttt{lukas.moddemann@hsu-hh.de} \\
	\And
	Daniel Vranješ \\
	Institute of Automation Technology\\
	Helmut Schmidt University\\
	Holstenhofweg 85, 22043 Hamburg, Germany \\
	\texttt{daniel.vranjes@hsu-hh.de} \\
	\And
	Oliver Niggemann \\
	Institute of Automation Technology\\
	Helmut Schmidt University\\
	Holstenhofweg 85, 22043 Hamburg, Germany \\
	\texttt{oliver.niggemann@hsu-hh.de} \\
}
\else
\usepackage{authblk}

\newbox{\orcid}\sbox{\orcid}{\includegraphics[scale=0.06]{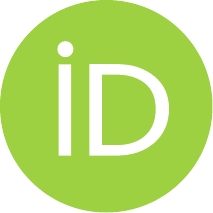}} 
\author[1]{Henrik Sebastian Steude}
\author[1]{Alexander Diedrich}
\author[2]{Ingo Pill}
\author[1]{Lukas Moddemann}
\author[1]{Daniel Vranješ}
\author[1]{Oliver Niggemann}
\affil[1]{Institute of Automation Technology, Helmut Schmidt University, Holstenhofweg 85, 22043 Hamburg, Germany}
\affil[2]{Institute of Software Technology, Graz University of Technology, Rechbauerstraße 12, 8010 Graz, Austria}
\fi

\renewcommand{\shorttitle}{Data-Driven Diagnosis for Large CPS with Minimal Prior Information}

\hypersetup{
pdftitle={Data-Driven Diagnosis for Large Cyber-Physical-Systems with Minimal Prior Information},
pdfsubject={cs.AI, eess.SY},
pdfauthor={Henrik Sebastian Steude, Alexander Diedrich, Ingo Pill, Lukas Moddemann, Daniel Vranješ, Oliver Niggemann},
pdfkeywords={Anomaly Detection, Diagnosis, Cyber-Physical Systems},
}

\begin{document}
\maketitle

\input{abstract}


\input{intro}

\input{sota}
\input{method}
\input{experiments}
\input{conclusion}
\section*{Acknowledgements}

This research paper is funded by dtec.bw---Digitalization and Technology Research
Center of the Bundeswehr. dtec.bw is funded by the European Union---NextGenerationEU.

\bibliographystyle{elsarticle-num}
\bibliography{references}

\end{document}

%% file: abstract.tex
\begin{abstract}
Diagnostic processes for complex cyber-physical systems often require extensive prior knowledge in the form of detailed system models or comprehensive training data.
However, obtaining such information poses a significant challenge. 
To address this issue, we present a new diagnostic approach that operates with minimal prior knowledge, requiring only a basic understanding of subsystem relationships and data from nominal operations.
Our method combines a neural network-based symptom generator, which employs subsystem-level anomaly detection, with a new graph diagnosis algorithm that leverages minimal causal relationship information between subsystems---information that is typically available in practice.
Our experiments with fully controllable simulated datasets show that our method includes the true causal component in its diagnosis set for 82\% of all cases while effectively reducing the search space in 73\% of the scenarios.
Additional tests on the real-world Secure Water Treatment dataset showcase the approach's potential for practical scenarios.
Our results thus highlight our approach's potential for practical applications with large and complex cyber-physical systems where limited prior knowledge is available.
\end{abstract}

%% file: intro.tex
\section{Introduction}
\label{sec:intro}

Deep learning methods have become recognized as state-of-the-art data-driven solutions for anomaly detection in Cyber-Physical Systems (CPSs) \cite{Garg2022-ol} due to their strong performance.
They allow us to learn models that capture the system behavior and to identify abnormal behavior as a deviation.
These approaches require nominal operation data only, and little to no additional prior knowledge.
However, identifying root causes of system failures represents a significantly more complex challenge than merely distinguishing between abnormal and nominal behavior \cite{Fink2020-tq}.
In order to address this challenge, existing approaches rely typically on methods such as supervised learning \cite{Chao-24-mpi}, classic symbolic AI techniques \cite{DIEDRICH2022104636} (including consistency-oriented~\cite{reiter1987theory,PQ15a} and abductive~\cite{MPW20} diagnosis), or hybrid methods \cite{Mohammadi2022-af, slimani2018fusion}.
All these concepts require precise prior structural knowledge, like detailed fault labels or formal system models, that are often hard to acquire.

The need for such detailed knowledge presents a significant practical barrier to developing diagnostic systems for large, complex CPSs like manufacturing plants or space stations.
Adding the required labels to historical sensor data is time-consuming, imprecise, and error-prone, particularly regarding the exact timing of events.
Furthermore, this approach limits the diagnostic capabilities to previously observed and labeled fault types, so we would potentially miss new or unforeseen failure modes~\cite{DIEDRICH2022104636}.
Creating precise and effective formal system models (or knowledge bases) for enabling symbolic consistency-oriented or abductive diagnostic reasoning, is likewise a complex, cumbersome, and often manual process, where we can draw only on limited tool support~\cite{MPW20}.
In the past, several open but crucial research challenges have thus been identified for this task~\cite{PdK24}.
For instance, structural analysis uses system models that consist mostly of well-defined differential equations \cite{frisk2017toolbox} and explicit fault models.
Similarly, case-based reasoning requires failure cases similar to the requirements of supervised learning approaches \cite{yan2024review}.

Addressing the shortcomings of available solutions, our primary motivation was to develop a new diagnostic methodology that minimizes the need for prior knowledge and manual intervention and that is applicable also to complex CPSs.
Given the vast number of sensor signals in modern systems (often with real-valued measurements), accurate anomaly detection is a crucial first step and prerequisite for effective diagnosis.

\begin{figure}[h]
	\centering
    \includegraphics[scale=1.0]{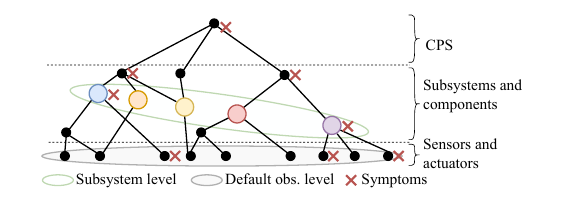}
        \caption{CPS system hierarchy illustrating subsystem and observation levels.
        Colored nodes correspond to subsystems in Fig.~\ref{fig:graph-intro}.
        Our approach models health states at the subsystem level, bridging the gap between whole-system multivariate and individual sensor univariate methods.}
	\label{fig:hierarchy-intro}
\end{figure}

This motivated our first \textbf{Research Question RQ1:} Can we generate inputs (symptoms) that are meaningful to diagnosis algorithms by enhancing data-driven anomaly detection methods with typically available prior knowledge? An intuitive motivating example is that we can usually divide CPSs into subsystems and modules as in Fig.~\ref{fig:hierarchy-intro}.
While this structural information has been exploited for enhancing anomaly detection through modular Neural Networks (NNs)~\cite{Steude2024,Agarwala2021-bk,Ehrhardt2024-hy,Vranjes2024-sg}, we are not aware of a systematic investigation about leveraging this information when applying deep learning for diagnostic purposes.

\begin{figure}[h]
	\centering
    \includegraphics[scale=1.0]{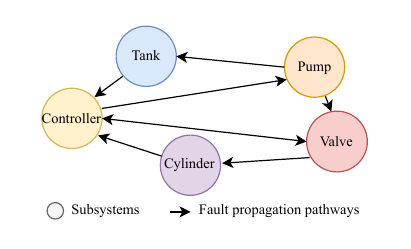}
        \caption{\textit{Causal subsystem graph} for a hydraulic system. Nodes represent subsystems and the edges encode fault propagation paths.}
	\label{fig:graph-intro}
\end{figure}

The desire to automate this diagnostic process is captured by \textbf{Research Question RQ2}: Can we identify the subsystems that caused the system failures using the symptoms from RQ1 without significant additional modeling efforts?
While data-driven models, as targeted by RQ1, can effectively detect anomalies and identify subsystems exhibiting abnormal behavior, they do not inherently solve the problem of pinpointing the subsystem(s) responsible for the system failure.
In addition to the system hierarchy, causal fault propagation graphs (see Fig.~\ref{fig:graph-intro}) are often available or can be derived with relatively little effort \cite{Tu2003-yu,weber2007diagnosing,Rehak2023-rd}. 
The purpose of such a graph is to capture the propagation of faults between subsystems.

For example, in an industrial hydraulic system, components like water tanks, pumps, valves, and hydraulic cylinders are interconnected.
A faulty pump might then cause abnormal pressure readings in connected valves, and a leaking tank could result in irregular flow patterns in downstream components like in Fig.~\ref{fig:graph-intro}.
In our experiments in Section~\ref{sec:experiments}, we thus aimed to explore how these graphs can be exploited to trace back symptoms observed in some subsystem and identify the subsystem with the original fault as root cause.

Motivated by RQ1 and RQ2, we developed the novel and comprehensive diagnostic approach introduced in the remainder of this paper.
During training, our solution requires only three inputs: \textit{(i)} observations from nominal system behavior, \textit{(ii)} a fault propagation graph, which we refer to as the \textit{causal subsystem graph}, and \textit{(iii)} the \textit{subsystem-signals map} as a mapping between signals and the subsystems they are connected to.
With this work, we make the following contributions:
\begin{itemize}
    \item A demonstration that CPS structure-informed deep learning models can detect symptoms in CPS subsystems and provide meaningful inputs for diagnostic processes.
    \item A novel diagnostic algorithm drawing on the \textit{causal subsystem graph} and detected symptoms.
    \item A systematic and incremental evaluation of each component of our methodology, drawing on simulated and real-world datasets.
\end{itemize}
In order to establish the context for our work and contributions, we provide a corresponding discussion of related work in the next section, followed by the presentation and evaluation of our diagnosis approach in the remainder of this paper.

%% file: sota.tex
\section{Related Work}
\label{sec:related_work}

Three research directions are particularly relevant for our study: anomaly detection, fault diagnosis, and causality research.
The following subsections explore various approaches within these areas and identify how they align with or diverge from our proposed methods.

\subsection{Anomaly Detection}
For multivariate time series, that are common in CPSs, deep learning models have demonstrated remarkable effectiveness and performance in detecting anomalies \cite{Pang2021-fv,Jeffrey2023-wx}, even for complex real-world scenarios~\cite{Garg2022-ol,Chen2022-qa}.
The corresponding approaches can be split into two primary categories: prediction and reconstruction. 
Prediction-based techniques often utilize modern neural architectures, including transformers \cite{Chen2022-qa}, temporal convolutional networks \cite{Cheng2019-sj}, and graph neural networks \cite{Deng2021-vd}.
Reconstruction-based approaches often integrate representation learning with frameworks such as generative adversarial networks \cite{Li2019-kd,Ciancarelli2023-ax} and variational autoencoders \cite{Chen2022-st,Lin2020-yi}.
While various studies have explored the nexus of anomaly detection and diagnostic processes \cite{Marino2021-yv,Garg2022-ol}, there remains a gap in integrating deep learning-based multivariate time series analysis with modern diagnostic methodologies.
This is particularly evident in the context of CPSs with limited prior knowledge.
Most existing approaches rely on well-labeled datasets which are often challenging to obtain in practical settings.

\subsection{Fault Diagnosis}
For multivariate time series that are common in CPSs, deep learning models have demonstrated remarkable effectiveness and performance in detecting anomalies \cite{Pang2021-fv,Jeffrey2023-wx}, even for complex real-world scenarios~\cite{Garg2022-ol,Chen2022-qa}.
From a research perspective, model-oriented fault diagnosis can be grouped into consistency-based approaches \cite{reiter1987theory}, abductive concepts \cite{MPW20}, structural approaches \cite{frisk2017toolbox}, and hybrid ones \cite{slimani2018fusion}.
Applying these methods to real-world CPSs requires high adaptation efforts and a significant amount of expert knowledge.
The increasing scale and complexity of modern CPSs for practical applications raise severe challenges~\cite{PdK24}.
Works like CatIO~\cite{MPW20} can help by automating some steps, and we saw also the advent of specific diagnosis methods for CPSs that rely on hybrid approaches combining structural and logical concepts~\cite{ignatiev2019model, matei2019hybrid, Mohammadi2022-af, slimani2018fusion}.
Still, the challenge remains that more scalable and adaptable diagnostic approaches are required to live up to the challenge of diagnosing complex real-world CPSs~\cite{PdK24}. 

\subsection{Causality and Fault Propagation}
Causality is the driving mechanism for fault propagation, and both are captured by our \textit{causal subsystem graph} as a key input to our approach. 
In particular, causality is understood as a propagation of values---as defined in the Qualitative Process Theory from Forbus \cite{forbus1984qualitative} and De Kleer \cite{de1984qualitative}.
Stumptner and Wotawa \cite{STUMPTNER20011} use causal graphs to diagnose tree-structured and acyclic systems.
Similar to our approach, Rehak et al. \cite{Rehak2023-rd} combine anomaly detection with a root cause analysis based on causal graphs.
However, their graph models individual measurements of the system alongside predefined root causes, and does not model subsystems.
Grünbaum et al. \cite{grunbaum2022quantitative} presented causal discovery methods from observations of CPSs, while Runge et al. \cite{runge2023causal} proposed a statistical causal inference method for time series data in the natural sciences.
More recently, Da Silva et al. \cite{da2024use} demonstrated how to exploit large language models to obtain causal information for CPSs.
Consequently, we can employ a variety of approaches for obtaining causal information, if it is unavailable.
For our approach, we assume the availability of suitable causal graphs such as those proposed by Weber and Wotawa~\cite{weber2007diagnosing} or Bozzano et al. \cite{bozzano2015smt}.

%% file: method.tex
\section{A Novel Diagnosis Approach}
\label{sec:methodology}
We'll start discussing our proposed approach by specifying the notation and formalizing the challenge we address.
Let's assume that a CPS is composed of a non-empty finite set of subsystems $S = \{s_1, ..., s_n\}$, with a finite and non-empty set of associated signals $P = \{p_1, ..., p_m\}$ where measurements are available for time steps $t_i$ in a finite time interval $T$.
We have time series data bound by $P$ and $T$.

Our approach draws on the following inputs:
\begin{enumerate}
    \item Training data that capture nominal behavior in a time series matrix \( \mathbf{X} \in \mathbb{R}^{|T| \times |P|} \) such that each row $i$ captures measurements at time point \( t_i \in T \), and each column $j$ corresponds to a specific sensor measurement \( p_j \in P \).
    \item A \textit{subsystem-signals map} \( \mathcal{M} : S \rightarrow 2^{P} \), which reports for each subsystem \(s_k \in S \) a corresponding set of signals \( P_s \subseteq P \) (as an element in the powerset \(\mathcal{P}(P)=2^{P}\)) that is relevant for monitoring $s_k$.
    \item A \textit{causal subsystem graph} \( G = (S, E) \), such that vertices represent subsystems, and directed edges \((s_l,s_m) \in E \subseteq S \times S \) that indicate that a fault in subsystem \( s_l \) can propagate to and affect subsystem \( s_m \).
\end{enumerate}

\begin{figure*}[h]
	\centering
	\includegraphics[width=\textwidth]{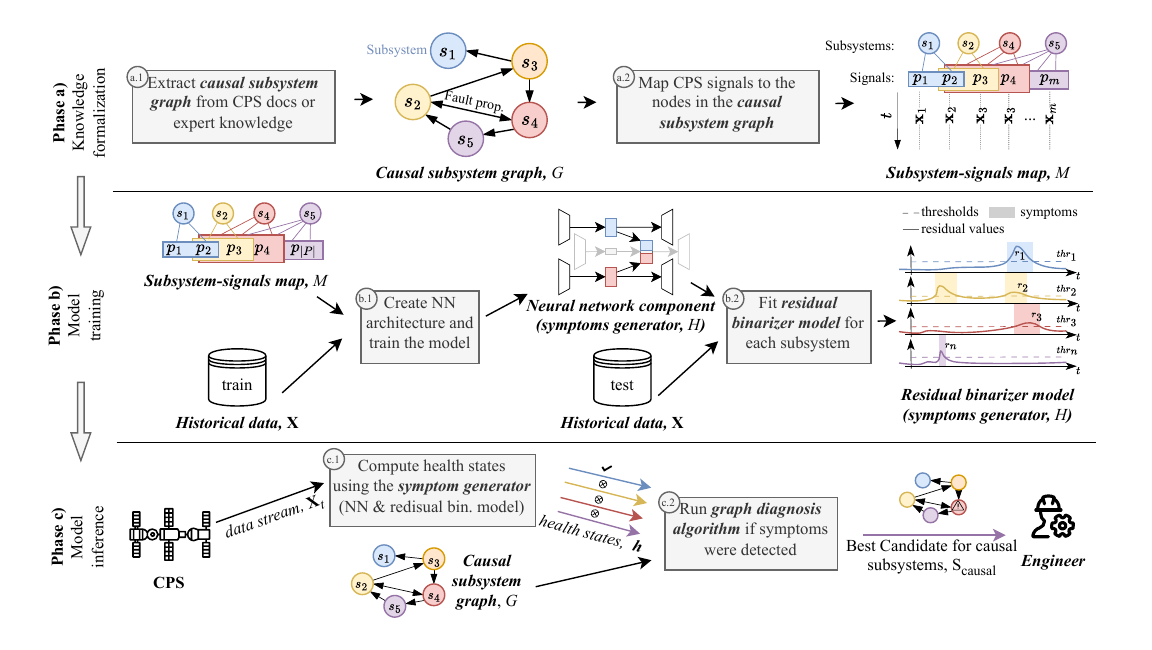}
    \caption{The three phases of our proposed diagnosis approach: a) knowledge formalization, b) model training, and c) model inference}
	\label{fig:flow-charts}
\end{figure*}

Based on these inputs, we define a diagnosis problem as follows: For a given time \( t \), we need to identify both the set of symptomatic subsystems \( S_{\text{sym}}(t) \subseteq S \) and the minimal set of subsystems \( S_{\text{causal}}(t) \subseteq S \) whose fault or abnormal behaviors are the root causes of all the observed symptoms in \( S_{\text{sym}}(t) \).
Formally, our approach aims to solve this problem by providing the following outputs:
\begin{enumerate}
    \item A set of symptomatic subsystems \( S_{\text{sym}}(t) \subseteq S \) and a health state vector \( \mathbf{h}(t) = (h_s(t))_{s \in S} \) where \( h_s(t) \in \{0, 1\} \), such that \( h_s(t) = 1 \) iff $s$ deviates from nominal behavior at $t \in T$ and $S_{\text{sym}}(t)$ contains exactly those subsystems for which we have \( h_s(t) = 1 \).
    \item A subset \( S_{\text{causal}}(t) \subseteq S \) containing those subsystems that are identified as the most probable root causes for all observed symptoms, based on the causal relationships encoded in graph \( G \).
\end{enumerate}

We now introduce the general concept of our approach as illustrated in Fig.~\ref{fig:flow-charts}.
It consists of three phases:

\textbf{Phase a) Knowledge formalization:}
Our approach is designed to work with minimal prior knowledge, requiring only the observational and minimal structural information described above.
As we argued in the introduction, the \textit{causal subsystem graph} \(G\) and \textit{subsystem-signals map} \(\mathcal{M}\) are often available from existing system documentation, or can easily be derived based on domain expertise. Therefore, we focus our technical descriptions and Algorithm \ref{alg:e2e-process} exclusively on the operational phases of model training and inference.

\textbf{Phase b) Model training:}
We utilize an NN architecture that is based on the \textit{subsystem-signals map} \(\mathcal{M}\) and was proposed in \cite{Steude2024}.
This approach enables anomaly detection at the subsystem level, an intermediate stage between individual sensors and the entire system (see Fig. \ref{fig:hierarchy-intro}).
We train this model with the time series data for nominal behavior described above (Step b.1 in Fig. \ref{fig:flow-charts}). 
The resulting residuals are subsequently processed by an additional model to generate the binary health state vector \(\mathbf{ h }\) for each \( t \in T \), such as to capture the operational status of each subsystem as either "OK" or "not OK" (Step b.2 in Fig. \ref{fig:flow-charts}). These components collectively form the \textit{symptoms generator} \( H \) that we describe from a more technical perspective in Sec.~\ref{subsec:symptoms-generator}.

\textbf{Phase c) Model Inference:}
During operations, we process the streamed data in two steps:
First, we apply the \textit{symptoms generator} once sufficient data is accumulated (Step c.1 in Fig. \ref{fig:flow-charts}) and generate health states for each subsystem that in turn form the health state vector \(\mathbf{h}(t)\).
Whenever the \textit{symptoms generator} detects anomalies, i.e., \(\exists s \in S: h_s(t) = 1\), we use the \textit{graph diagnosis algorithm} (Step c.2 in Fig. \ref{fig:flow-charts}) to analyze the causal relationships in \(G\) for identifying those nodes that are the most probable root causes of the observed symptoms.

\begin{algorithm}[h]
	\caption{Overall Diagnostic Process (Phases b and c only)}
	\label{alg:e2e-process}
	\begin{algorithmic}[1]
		\Require Training data $\mathbf{X}$, subsystem-signals map $\mathcal{M}$, causal subsystem graph $G$, window size $\Delta t$
		\Ensure Diagnostic results over time
		
		\State // Phase b) Model Training
		\State $H \gets \textsc{TrainSymptGen}(\mathbf{X}, \mathcal{M})$ \Comment{Section \ref{subsec:symptoms-generator}}
		
		\State // Phase c) Inference
		\While{system is running}
		\State // Step c.1: Apply the \textit{symptoms generator}
		\State $\mathbf{X}_t \gets \textsc{GetDataWindow}(\Delta t)$ \Comment{Latest sensor data}
		\State $\mathbf{h}(t) \gets H(\mathbf{X}_t)$ \Comment{Symptomatic subsystems}
		
		\State // Step c.2: Apply the \textit{graph diagnosis algorithm}
		\If{symptoms detected in $\mathbf{h}(t)$}
		\State $S_{\text{causal}}(t) \gets \textsc{Diagnosis}(G, \mathbf{h}(t))$ \Comment{Section \ref{subsec:graph-diagnosis-algo}}
		\State \Return $S_{\text{causal}}(t)$ \Comment{Root cause subsystems}
		\EndIf
		\EndWhile
	\end{algorithmic}
\end{algorithm}

Having introduced the overall concept for our approach, we now provide more details on its two key components and our corresponding contributions: the \textit{symptoms generator} and the \textit{graph diagnosis algorithm}.

\subsection{The Symptoms Generator}
\label{subsec:symptoms-generator}

As outlined above and illustrated in Fig.~\ref{fig:flow-charts}, the \textit{symptoms generator} consists of two components: a neural network component for generating residuals, which was initially presented in \cite{Steude2024}, and a residual binarization component for deriving the health state vector.
Both components are explained in the following paragraphs.

\textbf{Neural Network Component}:
We use an NN for mapping the input matrix \( \mathbf{X}_t \) to a residual vector \( \mathbf{r}(t) \in \mathbb{R}^{|P|}\) such that each element of \( \mathbf{r}(t) \) quantifies the reconstruction error---typically measured as the mean squared error between the actual data \( \mathbf{X}_t \) and the reconstructed data \( \hat{\mathbf{X}}_t \).
These residuals serve as a measure of 'normality' for the modeled telemetry data window.
The architecture features a composite latent space structured to reflect the subsystem layout of a CPS, facilitating both failure isolation and the identification of cross-subsystem anomalies. 

\textbf{Residual Binarization:}
Garg et al. \cite{Garg2022-ol} discuss various techniques for binarizing residual values in time series anomaly detection models.
In our context, we cannot optimize thresholds based on labeled data, since we lack the corresponding information on the subsystem level.
In contrast, we employ heuristic approaches when establishing thresholds, i.e., we utilize sensor data from normal operation that were not used for training or hyper-parameter tuning.
For instance, a subsystem might be flagged as "not OK" if its residual exceeds two standard deviations above the mean of these non-training data points.
Using these simple heuristics proves to be particularly effective since the binarization process becomes straightforward when the NN consistently produces residuals that are distinctly different for "OK" and "not OK" conditions.

Once the health state vector \(\mathbf{h}(t)\) is derived via binarizing the residuals \(\mathbf{r}(t)\), we compile the corresponding set of symptomatic subsystems \(S_{\text{sym}}(t)\).

\subsection{A Novel Graph Diagnosis Algorithm}
\label{subsec:graph-diagnosis-algo}
Once the symptoms and in turn the symptomatic subsystems have been identified, we address the core diagnostic challenge: identifying the root causes of the observed failures. 
While this set distinguishes subsystems that merely suffer from propagated symptoms from actually faulty ones, the set may include both symptomatic and non-symptomatic subsystems.
The \textit{causal subsystem graph} \(G\) provides a lightweight and easily obtainable yet powerful form of prior knowledge for this task.
However, its utilization presents two key challenges, that have not been fully addressed by previous research:
First, \( G \) may include cyclic relationships and multiple interconnected paths.
For instance, Stumpner and Wotawa \cite{STUMPTNER20011} present an algorithm for a similar problem and require the graph to be acyclic.
The more recently proposed solution by Rehak et al. \cite{Rehak2023-rd} does utilize cyclic graphs, but addresses a different diagnostic problem.
In their graphs, a distinction is made between predefined root cause nodes and measurement nodes.
Second, as we aim to identify a subset \(S_{\text{causal}}(t)\) that explains all observed symptoms in \(S_{\text{sym}}(t)\), we must account for the possibility of multiple, independent root causes occurring simultaneously---like accounted for in the model-based diagnosis theory~\cite{reiter1987theory}.

In order to tackle these challenges, we propose a novel graph-based algorithm (see Algorithm \ref{alg:graph-diagnosis}) that incorporates fundamental engineering principles for fault diagnosis.
The algorithm evaluates potential root causes through multiple complementing criteria that reflect typical fault propagation patterns in physical systems:
\begin{enumerate}
    \item \textbf{Reachability:} Measures whether and how well a candidate node can reach the observed anomalies (line 23-27).
    \item \textbf{Distance:} Assesses the proximity of the candidate to dense clusters of anomalies in the graph, since nodes that are close to multiple anomalies are presumably more likely to be their common cause (lines 28-33).
    \item \textbf{Anomaly Status:} Prioritizes candidates that exhibit anomalous states themselves (lines 34-36).
    \item \textbf{Anomaly Chains:} Evaluates connected sequences of anomalous nodes that can be reached from the candidate, where longer chains of consecutive anomalies strengthen the likelihood of a causal relationship (lines 37-42).
\end{enumerate}

Based on these criteria, we compute for each candidate node a score as a weighted sum with weights \(\mathbf{w} = (w_1, w_2, w_3, w_4)\) such that \(w_i \in [0,1]\) and \(\sum_i w_i = 1\).
The maximum score among all candidates is denoted as \(\sigma_{\text{max}} \in (0,1]\).

Selecting individual weights allows us to adapt the algorithm to specific scenarios.
In linear production lines, for instance, failure propagation follows typically clear downstream paths, which increases the importance of the reachability criterion.
For a system with complex and cyclic interconnections like chemical plants or power grids, though, the anomaly chain aspect might be a favorable choice for a higher weight.

\input{./graph-algo.tex}

Our diagnostic algorithm (Algorithm~\ref{alg:graph-diagnosis}) implements a multi-step process that iteratively evaluates and refines potential candidates considering the scores:
\begin{enumerate}
    \item \textbf{Initialization:} Given graph $G = (S, E)$ and health state vector $\mathbf{h}$, we first identify all anomalous subsystems $S_{\text{sym}}$ as targets for our root cause analysis.
    \item \textbf{Candidate Identification:} The candidate set $C$ is computed as the exact set of nodes that can reach any of the anomalous subsystems (lines 1 and 2).
    \item \textbf{Candidate Evaluation:} We compute the score for each candidate $c \in C$ based on our four criteria (lines 7-11).
    \item \textbf{Root Cause Selection:} Candidates are selected as root causes if their score exceeds the product $\theta\sigma_{\text{max}}$, where $\sigma_{\text{max}}$ is the highest score among all remaining candidates and $\theta \in [0,1]$ is a parameter (lines 13-14).
    \item \textbf{Iterative Refinement:} If the current set of root causes does not explain \emph{all} anomalies (line 5), we iteratively add further ones (on the unexplained anomalies) until the set explains all anomalies (lines 15-18).
\end{enumerate}

Our approach does not exhaustively evaluate the entire diagnosis space, so that it is incomplete in a formal sense.
This is a deliberate design choice, though, that allows us to deal with the space explosion problem inherent to our context.
That is, for a graph with a set of nodes \( S \) and a set of edges \( E \), a complete variant would suffer from the worst-case time complexity of \(O(2^{|S|}(|S|+|E|))\)---since we would need to check each subset of nodes and verify its reachability to the symptoms.
For our algorithm, we chose a practical trade-off between completeness and computational efficiency, achieving a worst-case time complexity of \(O(|S|^2(|S|+|E|))\) (detailed derivation available in the code repository), where the quadratic factor comes from potentially needing to evaluate each node as a candidate in each iteration.
By adjusting the parameter \( \theta \) of Algorithm~\ref{alg:graph-diagnosis} we can control the corresponding sensitivity.
A lower threshold increases the number of potential root causes considered and results in a more exhaustive search, whereas a higher threshold concentrates on the most probable causes resulting in a more focused diagnostic process.

%% file: graph-algo.tex
\begin{algorithm}[h!]
\caption{Graph Diagnosis Algorithm}\label{alg:graph-diagnosis}
\begin{algorithmic}[1]
\Require Graph $G = (S,E)$, health state vector $\mathbf{h}$, weights $\mathbf{w}$, threshold $\theta$
\Ensure Set of root causes $S_{\text{causal}}$

\State $S_{\text{sym}} \gets \{s \in S \mid h_s = 1\}$ \Comment{Anomalous subsystems}
\State $C \gets \{v \in S \mid \exists s \in S_{\text{sym}}: \text{path}(v \rightarrow s)\}$ \Comment{Candidates}
\State $S_{\text{causal}} \gets \emptyset$ \Comment{Root causes}
\State $U \gets S_{\text{sym}}$ \Comment{Unexplained anomalies}

\While{$U \neq \emptyset \land C \neq \emptyset$}
    \For{$c \in C$}
        \State $\sigma_r(c) \gets \textsc{ComputeReachabilityScore}(c, U)$
        \State $\sigma_d(c) \gets \textsc{ComputeDistanceScore}(c, U)$
        \State $\sigma_a(c) \gets \textsc{ComputeAnomalyScore}(c, S_{\text{sym}})$
        \State $\sigma_c(c) \gets \textsc{ComputeChainScore}(c, S_{\text{sym}})$
        \State $\sigma(c) \gets w_1\sigma_r(c) + w_2\sigma_d(c) + w_3\sigma_a(c) + w_4\sigma_c(c)$
    \EndFor
    
    \State $\sigma_{\text{max}} \gets \max_{c \in C} \sigma(c)$
    \State $B \gets \{c \in C \mid \sigma(c) \geq \theta \cdot \sigma_{\text{max}}\}$
    
    \For{$b \in B$}
        \State $S_{\text{causal}} \gets S_{\text{causal}} \cup \{b\}$
        \State $U \gets U \setminus S_{\text{reach}}(b, U)$
        \State $C \gets C \setminus \{b\}$
    \EndFor
\EndWhile

\State \Return $S_{\text{causal}}$

\State
\State // Helper functions for score computation
\Function{ComputeReachabilityScore}{$c, U$}
    \State $S_{\text{reach}}(c, U) \gets \{u \in U \mid \exists \text{ path } c \rightarrow u\}$
    \State \Return $|S_{\text{reach}}(c, U)| / |U|$
\EndFunction
\Function{ComputeDistanceScore}{$c, U$}
    \State // Group anomalies by their distance from $c$
    \State $\text{shells}[d] \gets \{u \in U \mid \text{shortest\_path}(c, u) = d\}$
    \State // Max. density, normalizing by approx. surface $d^2$
    \State \Return $\max_{d} |\text{shells}[d]| / d^2$ 
\EndFunction
\Function{ComputeAnomalyScore}{$c, S_{\text{sym}}$}
    \State \Return $\mathds{1}_{S_{\text{sym}}}(c)$ // Indicator function
\EndFunction
\Function{ComputeChainScore}{$c, S_{\text{sym}}$}
    \State // Let $\text{paths}(c)$ be all directed paths starting at $c$
    \State // Find longest path through symptomatic nodes only
    \State $\text{max\_length} \gets \max\{|p| : p \in \text{paths}(c) \land p \subseteq S_{\text{sym}}\}$
    \State \Return $\text{max\_length} / |S_{\text{sym}}|$ 
\EndFunction
\end{algorithmic}
\end{algorithm}

%% file: experiments.tex
\section{Experiments}
\label{sec:experiments}

For a systematic evaluation of our method's components and in order to address RQ1 and RQ2, we designed experiments that explore three hypotheses:
\begin{enumerate}
    \item[\textbf{H1}] The \textit{graph diagnosis algorithm} identifies causal subsystems \(S_{\text{causal}}\) reliably, assuming that the symptoms are accurately detected and \(G\) captures the system's causal relationships adequately.
    \item[\textbf{H2}] Given accurate \(G\) and \(\mathcal{M}\), our method effectively identifies the causal subsystems.
    \item[\textbf{H3}] The concept, methodology, and implementation of our diagnostic approach are viable and effective at diagnosing a real-world CPS.
\end{enumerate}
The code for all our experiments (along with documentation and additional details) is publicly available\footnote{\url{https://github.com/hsteude/ad-diag-end2end-experiments}} under the MIT license.
We refer the interested reader to \cite{Steude2024} for an external empirical validation of the symptoms generator.

\subsection{Experiment 1: The Graph Diagnosis Algorithm}
\label{subsec:experiment1}
This experiment was designed to validate \textbf{H1}, assuming that \(G\) represents the actual causal relationships and \(\mathbf{h}\) correctly indicates symptoms.
To test the algorithm's performance across a range of conditions, we defined four scenarios that vary in their symptom distribution complexity and graph structure.
For a balanced evaluation across these diverse scenarios, we employ equal weights (\(w_1 = w_2 = w_3 = w_4 = 0.25\)) for all evaluation criteria.
The four scenarios are:
\begin{itemize}
    \item \textbf{Scenario 1:} An acyclic graph with a single symptom cluster as a baseline case for causal analysis.
    \item \textbf{Scenario 2:} An acyclic graph containing two independent symptom clusters for evaluating multi-fault capabilities.
    \item \textbf{Scenario 3:} A cyclic graph with a single symptom cluster as baseline for assessing the impact of feedback loops.
    \item \textbf{Scenario 4:} A cyclic graph with multiple symptom clusters for investigating the multi-fault performance in the context of cyclic dependencies.
\end{itemize}

\begin{figure}[h]
    \centering
    \includegraphics[scale=0.63]{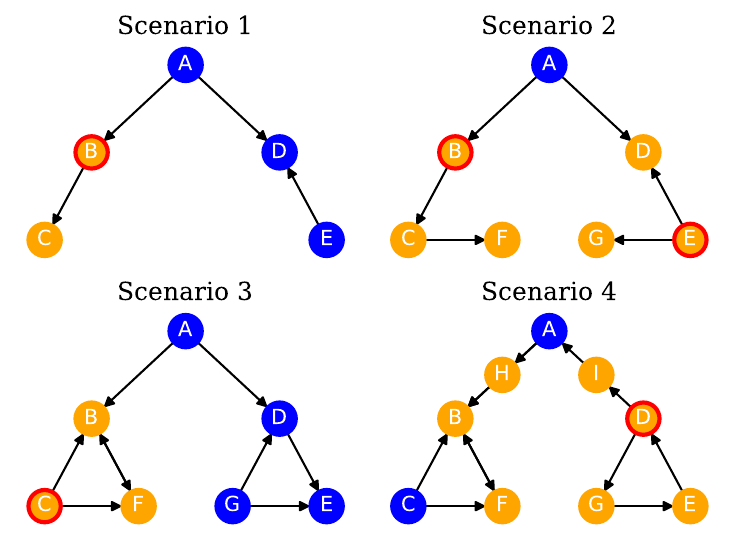}
    \caption{Exp.~1 with \(\theta = 1.0\). Nominal subsystems are blue, yellow ones indicate detected symptoms and red circles highlight root causes.}
    \label{fig:results-exp1}
\end{figure}

Fig.~\ref{fig:results-exp1} illustrates the four scenarios and also the predictions of the \textit{graph diagnosis algorithm}.
The results demonstrate the algorithm's effectiveness across varying levels of graph complexity.
In Scenarios 1 and 2, the algorithm identified both single and multiple root causes in acyclic graphs successfully, aligning with intuitive causal reasoning. 
This capability extends to Scenario 3, where the algorithm maintains its effectiveness despite the presence of cyclic dependencies.
Scenario 4, the most complex case, reveals two key insights:
First, fault propagation is indicated through node A, which appears healthy, reflecting the algorithm's design to account for symptom-free propagation paths.
Second, when setting \(\theta = 1.0\), only node "D" is identified as potential root cause, due to its higher distance and chain metric scores.
Nodes "E" and "G" are included in the diagnosis when \(\theta\) is reduced, demonstrating the parameter's role in controlling the algorithm's sensitivity.

In order to analyze this behavior systematically, we varied \(\theta\) from 1.0 to 0.0 in 0.1 decrements.
Table \ref{tab:theta-sensitivity} shows how nodes are progressively included in the diagnosis as \(\theta\) decreases, with \(theta=1.0\) corresponding to the results in Figure \ref{fig:results-exp1}.

\begin{table}[h]
\centering
\caption{Sensitivity analysis for parameter \(\theta\). We report only thresholds resulting in new nodes.}
\label{tab:theta-sensitivity}
\begin{tabular}{c|l|l|l}
\textbf{Scenario} & \textbf{\(\theta\)} & \textbf{All Root Causes} & \textbf{Newly Added} \\ \hline
\multirow{3}{*}{1} & 1.00 & B & B \\
 & 0.80 & A, B & A \\
 & 0.60 & A, B, C & C \\ \hline
\multirow{4}{*}{2} & 1.00 & B, E & B, E \\
 & 0.80 & A, B, E & A \\
 & 0.70 & A, B, C, E & C \\
 & 0.50 & A, B, C, D, E, F, G & D, F, G \\ \hline
\multirow{3}{*}{3} & 1.00 & C & C \\
 & 0.70 & B, C, F & B, F \\
 & 0.50 & A, B, C, F & A \\ \hline
\multirow{6}{*}{4} & 1.00 & D & D \\
 & 0.90 & D, G & G \\
 & 0.80 & D, E, G & E \\
 & 0.70 & D, E, G, H & H \\
 & 0.60 & A, B, D, E, F, G, H & A, B, F \\
 & 0.50 & A, B, C, D, E, F, G, H, I & C, I \\
\end{tabular}
\end{table}

In summary, the algorithm effectively identifies causal subsystems in both cyclic and acyclic graphs, aligning with intuitive classifications.
The sensitivity analysis shows how lowering \( \theta \) progressively includes more potential root causes, based on decreasing scores from our evaluation criteria.
Although our approach is not complete in the sense that it does not consider every possible combination of candidate nodes, by progressively lowering the threshold, each candidate is eventually included in the set of predicted root causes.

\subsection{Experiment 2: Controlled Simulated Datasets}
\label{subsec:experiment2}
In order to validate \textbf{H2}, we used simulated datasets where both \(G\) and \(\mathcal{M}\) are known to be accurate.
This approach allows for controlled manipulation of dataset characteristics, enabling a thorough assessment of our methodology.

\begin{figure*}[h]
    \centering
    \includegraphics[width=\textwidth]{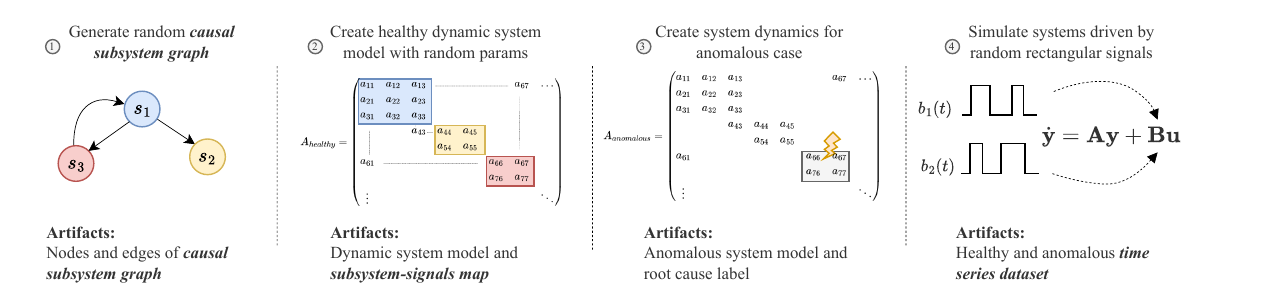}
    \caption{Illustration of the time series data generation process, showing four key stages. Each matrix entry, such as \(a_{11}\), \(a_{44}\), is randomized; unspecified entries are assumed to be zero, indicating no direct influence between those states.}
    \label{fig:data-gen-process}
\end{figure*}

Each trial involves generating a random graph \(G\) and modeling system dynamics with a matrix \(\mathbf{A}\) (Figure \ref{fig:data-gen-process}).
The state dynamics follow \(\dot{\mathbf{y}} = \mathbf{A}\mathbf{y} + \mathbf{B}\mathbf{u}\), where \(\mathbf{y}\) is the state vector and \(\mathbf{u}\) the control inputs.
Nominal operation data is generated by applying random control signals, while abnormal conditions are created by altering specific parameters in \(\mathbf{A}\).
We divide the nominal operation data into training and validation sets for the \textit{symptoms generator}, reserving an additional dataset for setting thresholds of the \textit{residual binarization model}.
For this purpose, we employ a straightforward threshold-based approach: we compare the median of residuals from abnormal periods to the 75th percentile of the nominal test set, as the simulated data provides clearly separated nominal and abnormal operation periods.
The \textit{symptoms generator} then identifies anomalies, which the \textit{graph diagnosis algorithm} analyzes.

We conducted 100 trials with varying graph sizes (5-100 nodes), edge densities, noise levels, and anomaly intensities.
Each node represents 2-5 signals, resulting in datasets of dozens to hundreds of signals.
We categorized the trials into five groups:
\textit{(i)} Trials where the causal subsystem \( s_{\text{true}} \in S \) is not classified as anomalous by the \textit{symptoms generator}, i.e. \( s_{\text{true}} \notin S_{\text{sym}} \).
\textit{(ii)} Trials where the causal subsystem is detected as anomalous but not identified as a root cause in the diagnosis, i.e. \( s_{\text{true}} \in S_{\text{sym}} \), but \( s_{\text{true}} \notin S_{\text{causal}} \)
\textit{(iii)} Trials where the causal subsystem is correctly identified but the diagnosis fails to reduce the number of candidates compared to the symptoms, i.e. \( s_{\text{true}} \in S_{\text{sym}} \) and \( s_{\text{true}} \in S_{\text{causal}} \) but \(|S_{\text{causal}}| \geq |S_{\text{sym}}|\).
\textit{(iv)} Trials where the diagnosis successfully reduces the candidate set but still includes multiple subsystems including the true cause, i.e. \( s_{\text{true}} \in S_{\text{causal}} \) and \(|S_{\text{causal}}| < |S_{\text{sym}}|\) but \(|S_{\text{causal}}| > 1\).
\textit{(v)} Trials with perfect diagnosis where the true causal subsystem is identified as the sole root cause, i.e. \(S_{\text{causal}} = \{s_{\text{true}}\}\).
We set \(\theta=0.9\) for all trials to balance precision and inclusivity.

\begin{figure}[h]
    \centering
    \includegraphics[width=0.7\textwidth]{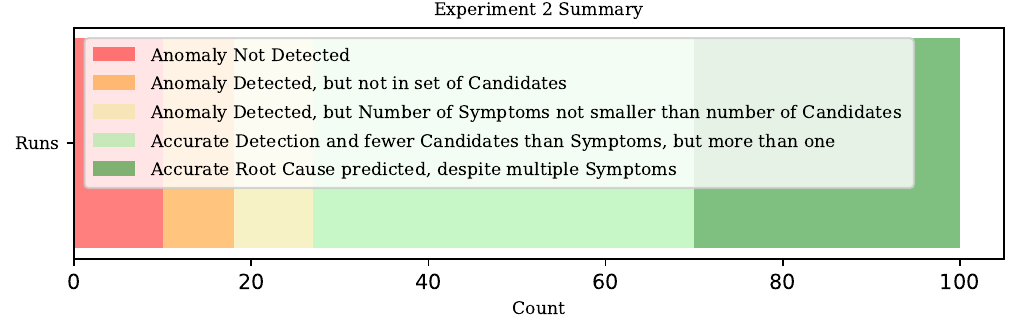}
    \caption{Summary of Experiment 2 outcomes over 100 trials.}
    \label{fig:exp2-result-bar-chart}
\end{figure}

Figure \ref{fig:exp2-result-bar-chart} summarizes the outcomes across these categories.
The true causal subsystem is included in the predicted root causes in 82\% of trials.
In 73\% of cases, our method reduces the search space compared to the initial symptoms detection.
Notably, 30\% of trials achieve perfect diagnosis, identifying the true causal subsystem as the sole root cause.
The remaining cases include scenarios where the causal subsystem is missed in the final diagnosis (8\%) or not detected as anomalous initially (10\%).
These cases occur primarily with subtle parameter alterations during data generation and might be addressed through refined threshold optimization.
For all trials, we set \(\theta=0.9\) to maintain a balance between diagnostic precision and the risk of excluding potential root causes, as indicated by the sensitivity analysis in Table \ref{tab:theta-sensitivity}.

Overall, these results demonstrate our method's effectiveness in identifying root causes, even with complex cyclic dependencies and large multivariate datasets.

\subsection{Experiment 3: The SWAT Dataset}
\label{subsec:experiment3}

In this subsection, we detail our experiment applying the proposed diagnostic approach to the Secure Water Treatment (SWAT) dataset \cite{Goh2017-mr}, a real-world dataset often used in research on industrial control systems.
While public datasets that satisfy all our requirements are scarce, the SWAT dataset fulfills two essential criteria: first, it is well-documented, allowing the construction of \(G\) and \(\mathcal{M}\), and second, it includes detailed fault information, facilitating the validation of fault diagnosis methodologies.
The SWAT dataset encompasses data from nominal operations and specific induced faults across six stages of a water treatment process, each monitored by a network of sensors and actuators, generating 51 distinct signals, which makes the system relatively small compared to the ones we used in Experiment 2.

\begin{figure}[h]
    \centering
    \includegraphics[width=0.7\textwidth]{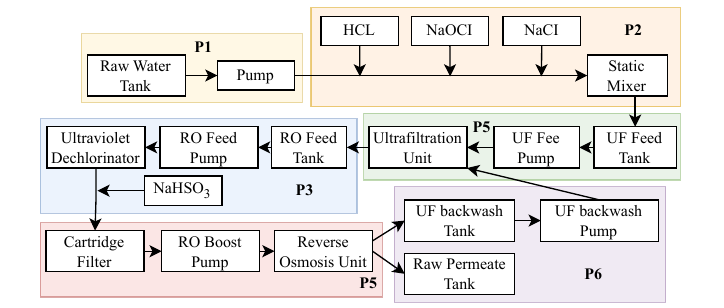}
    \caption{SWAT process diagram, as visualized in \cite{Goh2017-mr}.}
    \label{fig:swat-docu-graph}
\end{figure}

The setup of our experiment involves executing the three phases of the process depicted in Fig. \ref{fig:flow-charts}.
For \textbf{knowledge formalization} (Phase a), we employ the documentation \cite{Goh2017-mr} to create \(G\) and \(\mathcal{M}\).
We define subsystem boundaries corresponding to process steps P1 through P6, as illustrated in Figure \ref{fig:swat-docu-graph}.
The assignment of signals to these subsystems is automated based on their nomenclature: each signal label includes a three-digit numeric code, where the first digit indicates the process step, thereby identifying the subsystem.

The main steps of \textbf{model training} (Phase b) follow the process described in Experiment 2, with the key difference being that data are directly imported from the SWAT dataset rather than generated.
Our experimental pipeline incorporates minor data cleaning and preprocessing steps, detailed in our code repository.
For the \textit{residual binarization model}, we smooth the residuals using a moving median.
Due to higher volatility in the normal operation residuals, we employ a more conservative threshold (99th percentile) compared to Experiment 2, calculated from anomaly-free test data.

For \textbf{model inference} (Phase c), we initially filter the list of documented attacks in the SWAT dataset to focus on those with a defined impact and predictable outcomes, creating a set of anomalies for which the 'attack point' can be assumed to be within the causal subsystem \(s_{\text{causal}}\).
For the graph diagnosis algorithm, we use weights of \(\mathbf{w} = (0.2, 0.2, 0.4, 0.2)\) for the evaluation criteria, placing greater emphasis on the anomaly status.
To validate our methodology, we systematically apply our comprehensive method to each identified attack.

\begin{table}[!ht]
\centering
\caption{Diagnostic results on the SWAT dataset.}
\begin{tabular}{c|c|c|c}
\textbf{Attack} & \textbf{Attacked Subsys.} & \textbf{Symptoms} & \textbf{DX Candidates} \\ \hline\hline
    \#01 & \textbf{P1} & \textbf{P1, P5} & \textbf{P1} \\ \hline
    \#02 & P1 & P1, P5, P6 & P5, P1 \\ \hline
    \#17 & \textit{P3} & \textit{P1, P3} & \textit{P1} \\ \hline
    \#21 & P1 & P1, P5 & P1 \\ \hline
    \#23 & P6, P3 & P3 & P3 \\ \hline
    \#25 & \textit{P4} & - & - \\ \hline
    \#26 & P1, P3 & P1, P6 & P1 \\ \hline
    \#27 & \textit{P3, P4} & \textit{P2} & \textit{P2} \\ \hline
    \#28 & \textbf{P3} & \textbf{P3, P4, P5} & \textbf{P3} \\ \hline
    \#30 & P1, P2 & P1 & P1 \\ \hline
    \#35 & \textbf{P1} & \textbf{P1, P2, P6} & \textbf{P1} \\
\end{tabular}
\label{tab:results-exp3}
\end{table}

The efficacy of the \textit{symptom generator} in identifying anomalies across the system was previously demonstrated on the SWAT dataset in \cite{Steude2024}.
We thus focus on the diagnosis results in our analysis.
Table \ref{tab:results-exp3} illustrates the results of our comprehensive diagnostic method regarding the analysis of fault cases.
In eight out of eleven cases, at least one attacked subsystem was included in the set of predicted root causes.
Notably, in three cases (bold rows), the algorithm narrowed multiple symptoms to a single affected subsystem.
However, the italicized rows highlight four cases where the attack points did not correspond to the subsystems flagged as abnormal by the symptom generator, or were not selected as rootcause candidate by the graph algorithm.

Our method shows promise in identifying subsystems involved in anomalies, with attacked subsystems included in the predicted root causes in most cases.
However, these results rely on assumptions about system causality that we could not fully validate due to limited system understanding or incomplete documentation.

%% file: conclusion.tex
\section{Conclusion}

This paper introduces a novel diagnostic approach that effectively integrates deep learning-based anomaly detection with a graph-based diagnostic approach in order to address the complexities of diagnosing complex CPSs with limited prior knowledge.
Our approach leverages the strengths of both data-driven and symbolic AI techniques, i.e., we exploit \textit{causal subsystem graphs} and \textit{subsystem-signals maps} that in their union facilitate the interpretation of system interactions in terms of causality and the context of symptoms.

We conducted experiments on fully controllable simulated datasets and the real-world SWAT dataset.
For the synthetic scenarios, we identified the true causal component correctly in 82\% of the cases and reduced the search space for root causes for 73\%. 
Similarly, when applied to the SWAT dataset, our method successfully included attacked subsystems in the predicted root causes in eight out of eleven cases.
These results demonstrate that our approach effectively identifies root causes across both synthetic and real-world industrial scenarios.

That said, we would like to remind the reader that this effectiveness depends to a large extent on the quality and accuracy of the causal subsystem graph and the representativeness of the nominal training data. Some further implicit assumption is the observability of subsystems through signals.
Future work will thus have to investigate the effects of suboptimal input data as well as observability issues to the effect of being able to automatically detect those and in turn derive, e.g., sensor placement strategies for optimizing a system's diagnosability.
A direction for further future research could be the evaluation of other graph formats in terms of their effects on the achievable diagnostic performance.